\documentclass{article}

\usepackage[final]{neurips_2024}

\usepackage[utf8]{inputenc} 
\usepackage[T1]{fontenc}    
\usepackage{hyperref}       
\usepackage{url}            
\usepackage{booktabs}       
\usepackage{amsfonts}       
\usepackage{nicefrac}       
\usepackage{microtype}      
\usepackage{xcolor}         

\usepackage{subcaption} 
\usepackage{graphicx} %
\usepackage{bm}
\usepackage{amsmath}
\usepackage{amssymb}
\usepackage{mathtools}
\usepackage{amsthm}
\usepackage{enumitem}

\newtheorem{thm}{Theorem}
\newcommand{\aref}[1]{\hyperref[#1]{Appendix~\ref*{#1}}}
\newcommand{\figrefA}[1]{\hyperref[#1]{Figure~\ref*{#1}A}}
\newcommand{\figrefB}[1]{\hyperref[#1]{Figure~\ref*{#1}B}}
\newcommand{\figrefC}[1]{\hyperref[#1]{Figure~\ref*{#1}C}}
\newcommand{\figrefBC}[1]{\hyperref[#1]{Figure~\ref*{#1}B\&C}}

\title{Learned Random Label Predictions as a Neural Network Complexity Metric}

\author{%
  Marlon Becker\\
  University of Münster, Germany\\
  \texttt{marlon.becker@uni-muenster.de} \\
  \And
  Benjamin Risse\\
  University of Münster, Germany\\
  \texttt{b.risse@uni-muenster.de} \\
}

\begin{document}

\maketitle

\begin{abstract}
  We empirically investigate the impact of learning randomly generated labels in parallel to class labels in supervised learning on memorization, model complexity, and generalization in deep neural networks.
  To this end, we introduce a multi-head network architecture as an extension of standard CNN architectures.
  Inspired by methods used in fair AI, our approach allows for the unlearning of random labels, preventing the network from memorizing individual samples.
  Based on the concept of Rademacher complexity, we first use our proposed method as a complexity metric to analyze the effects of common regularization techniques and challenge the traditional understanding of feature extraction and classification in CNNs.
  Second, we propose a novel regularizer that effectively reduces sample memorization.
  However, contrary to the predictions of classical statistical learning theory, we do not observe improvements in generalization.
\end{abstract}

\section{Introduction}

Modern deep learning models are highly prone to overfitting due to their dramatic overparameterization \citep{doubleDescent}.
These models not only overfit under standard training conditions, but can also achieve 100\% training accuracy on datasets with randomly generated labels \citep{zhang_understanding_2021}.
This demonstrates that modern ANNs are capable of memorizing sample-specific information, which is not class-related and therefore irrelevant to the desired task.
In essence, overfitting occurs in the most literal sense, i.e., by fitting each individual sample in the training data.\\
Beyond the intuitive understanding of a model's ability to fully memorize training data, its performance on randomly generated labels offers a valuable complexity measure within the theoretical framework of PAC learning.
Specifically, Rademacher complexity measures the capacity of a binary classifier by evaluating how well an optimal model from a given hypothesis class can fit random labels.
Although finding the optimal model is not guaranteed for SOTA neural networks, training with SGD on random labels provides an empirical estimate of Rademacher complexity.
Reducing a model's Rademacher complexity is closely tied to improved generalization bounds (see \aref{sec:rademacher}).
The primary objective of this work is to assess random label prediction accuracy as a complexity metric and to develop a random label regularizer, which, according to classical learning theory, is predicted to enhance generalization.\\
While it is straightforward to assess a model's memorization capacity by training it on random labels, effect on generalization for a given task as well as regularization can only be applied when random labels are learned simultaneously during training on correctly labeled data.
This paper addresses these challenges through the following contributions:
\begin{itemize}[noitemsep,topsep=0pt]
  \item We propose a novel neural network architecture built on top of classical CNNs that learns class labels and randomly generated labels in parallel, allowing for regularization of random labels without compromising the primary objective of class learning.
  \item We evaluate our proposed memorization metric in the context of other regularization techniques known to affect model complexity, namely weight decay, dropout, and label smoothing.
  \item We utilize the memorization metric to challenge the traditional understanding of the transition from feature extraction to classification in CNNs.
  \item While our results show that our proposed regularization effectively reduces memorization, we do not observe improved generalization, raising questions about the direct causal relationship between memorization and generalization as suggested by classical learning theory.
\end{itemize}

\section{Methods}

\begin{figure}[htb]
  \centering
  \begin{subfigure}[b]{0.33\textwidth}
    \centering
      \includegraphics[width=1\textwidth]{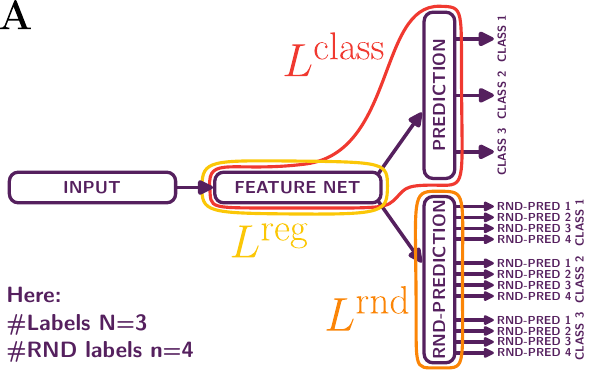}
  \end{subfigure}%
  \hspace{0.5cm}
  \begin{subfigure}[b]{0.33\textwidth}
    \centering
    \includegraphics[width=1\textwidth]{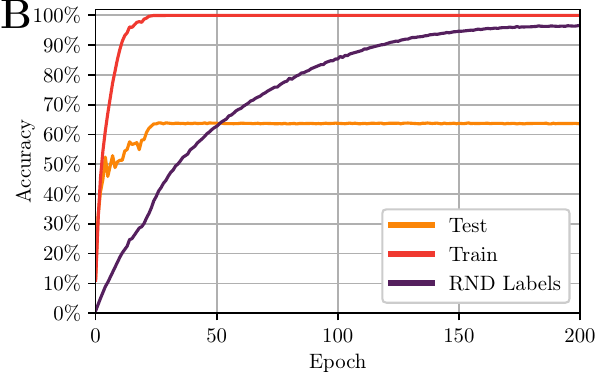}
  \end{subfigure}
  \caption{
    \textbf{A:} Illustration of the multi-head architecture that is built on top of the feature extractor to predict random labels.
    \textbf{B:} While test and train accuracy converge quickly, the random label accuracy still increases after reaching nearly $100\%$ train accuracy, and finally also reaches close to $100\%$.
  }
  \label{fig:RLR:scheme__learning_curve__copy_depth}
\end{figure}

Instead of training the network only on random labels, as done in previous work, we introduce additional heads to predict both the randomly generated label $s$ and the class label $y$.
These additional heads are constructed per class to predict the random label.
This setup produces predictions for all combinations of random labels ($n$ possible values) and class labels ($N$ possible values).
While traditional architectures produce a prediction vector $p \in \mathbb{R}^N$, our proposed architecture adds a second output $\hat{p} \in \mathbb{R}^{N \times n}$.
A visualization of this architecture is shown in \figrefA{fig:RLR:scheme__learning_curve__copy_depth}.
Although a single additional head would suffice to predict the random labels, we chose a nested structure to allow for a regularization loss that does not conflict with the class prediction objective, as discussed below.
This scheme is inspired by fair AI, where unwanted biases present in the training data are suppressed without deteriorating the actual learning objective.
Three losses are used to train the different parts of the model.
Using the Kronecker delta $\delta_{ij}$, these are:
\begin{equation}
  L^{class} = -\sum_{i=1}^N \delta_{iy} \log(p_i) \qquad L^{rnd} = -\sum_{j=1}^{N} \delta_{jy} \sum_{i=1}^{n} \delta_{is}  \log(\hat{p}_i^j) \qquad L^{reg} = - \sum_{j=1}^{N} \delta_{jy} \frac{1}{n} \sum_{i=1}^{n} \log(\hat{p}_i^j)\nonumber
\end{equation}
The classification loss $L^{class}$ is the cross-entropy between the one-hot encoding of the class labels $y$ and the output of the main classification head $p$.
As in standard training settings, this loss is used to train the full baseline model, including both the feature extractor and the default classification head.
Second, the random prediction heads are trained using $L^{rnd}$, which selects the output of the random prediction heads corresponding to the correct class label $y$ \textit{and} correct random label $s$. 
A third loss term is introduced to regularize the feature extractor by unlearning the random labels.
However, simply reversing the sign of the random label loss $L^{rnd}$ leads to unstable training, since the simplest strategy to minimize such a loss is to learn nothing at all.
To address this, we introduced the mentioned separate heads for each class, which allows for a non-contradictory formulation of the regularization loss.
The regularization loss $L^{reg}$ seeks to equalize the probabilities for all random labels while ensuring accurate class label predictions.
The loss is set to the cross-entropy between a uniform distribution and the predictions of the random prediction head corresponding only to the correct class.
This regularization term, scaled by a tunable hyperparameter $\lambda$, is added to the loss of the feature extractor.
The random prediction head is trained to capture sample-specific information from the feature extractor, since random label predictions can only be correct if individual input samples are memorized.
The regularizer acts as an adversary to this by encouraging the features to abstract away from individual samples, thereby preventing accurate random label predictions while still retaining class information for correct class label predictions.

\section{Experiments}
Unless stated otherwise, the following experiments are performed for a WideResNet-16-4 \citep{WRN} on CIFAR100 \citep{CIFAR} with SGD with momentum $\mu=0.9$, a cosine learning rate scheduler with a base learning rate of $\eta = 0.5$ and a batch size of 256 trained for 200 epochs without additional regularization and the number of random labels is set to $n=10$.

\subsection{Complexity Metric}
The accuracy of the random prediction heads can be used to define a complexity metric that serves as an empirical approximation to Rademacher complexity.
When the regularizer is not applied (i.e., $\lambda = 0$), the random prediction heads do not affect the baseline network.
Instead, they act only as a metric.

\subsubsection{Classification Features are Sufficient for Memorization}
The random prediction heads, consisting of only a single fully connected layer, are sufficient to achieve high accuracies on the randomly generated labels, as shown in the learning curves in \figrefB{fig:RLR:scheme__learning_curve__copy_depth}.
This phenomenon was previously observed only for full multi-layer networks trained on the random prediction task before~\citep{zhang_understanding_2017}.
Since the feature extraction part of the network is not trained to predict the random label, sample-specific information must be present in the feature output. 
Notably, while the test and training accuracies converge rapidly, the accuracy of the random predictions continues to increase even after the training accuracy approaches 100\%.
This could indicate increasing overfitting to the training set even after a perfect training accuracy is achieved.

\subsubsection{Common Regularizers Reduce Memorization}
\begin{figure}[htb]
  \centering
  \begin{subfigure}[b]{0.33\textwidth}
    \centering
    \includegraphics[width=1\textwidth]{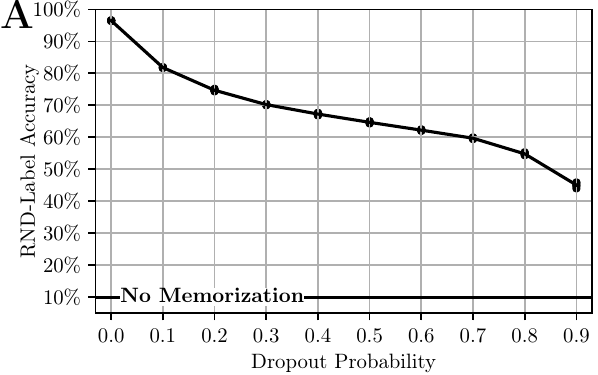}
  \end{subfigure}%
  \begin{subfigure}[b]{0.33\textwidth}
    \centering
    \includegraphics[width=1\textwidth]{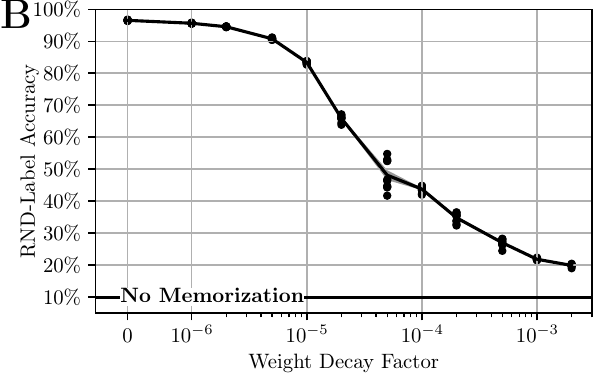}
  \end{subfigure}%
  \begin{subfigure}[b]{0.33\textwidth}
    \centering
    \includegraphics[width=\textwidth]{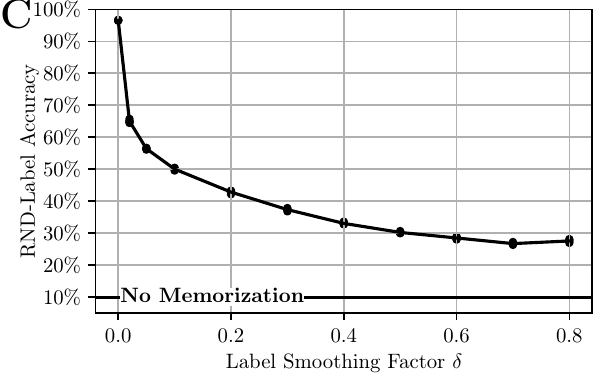}
  \end{subfigure}
		\caption{
      The effect of common complexity regularizers can be measured with the proposed metric.
      \textbf{A:} Dropout.
      \textbf{B:} Weight decay.
      \textbf{C:} Label smoothing.
      }
      \label{fig:other_regs}
\end{figure}

We confirm that the accuracy of the random label predictions correlates with model complexity by evaluating our proposed metric for three well-known regularization techniques: dropout, weight decay, and label smoothing.
Each of these regularizers effectively reduces the random label accuracy, as shown in \autoref{fig:other_regs}.

\subsubsection{Feature Extraction vs. Classification: Copy Depth}
\begin{figure}[htb]
  \centering
  \begin{subfigure}[b]{0.33\textwidth}
    \centering
    \includegraphics[width=1\textwidth]{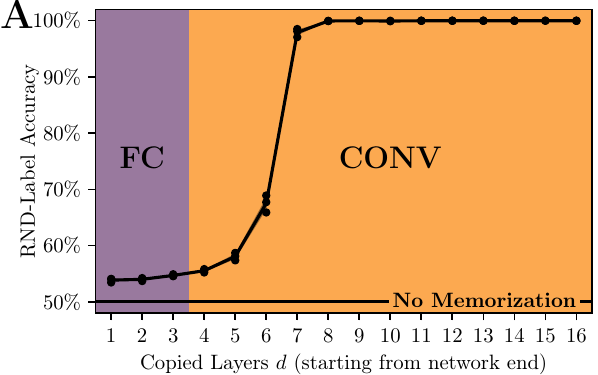}
  \end{subfigure}%
  \begin{subfigure}[b]{0.33\textwidth}
    \centering
    \includegraphics[width=1\textwidth]{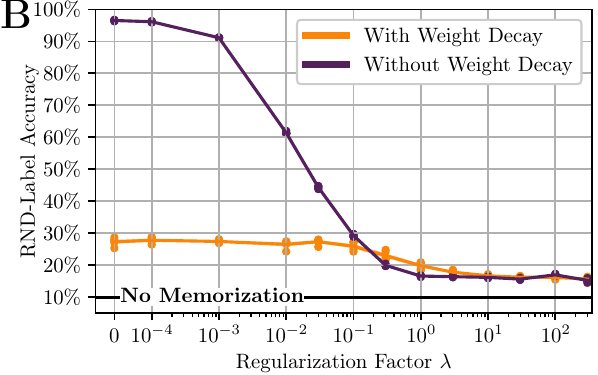}
  \end{subfigure}%
  \begin{subfigure}[b]{0.33\textwidth}
    \centering
    \includegraphics[width=1\textwidth]{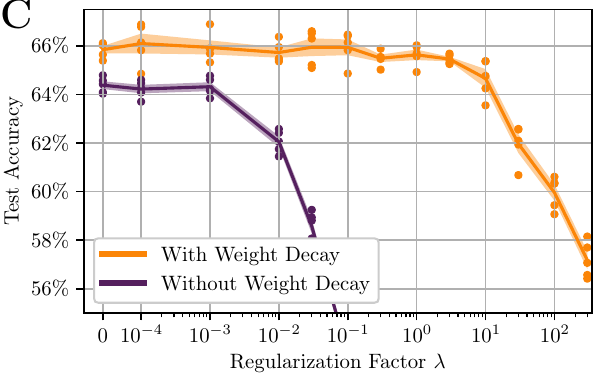}
  \end{subfigure}
  \caption{
    \textbf{A:} Effect of the number of copied layers, i.e., the copy depth $d$, on the random label prediction accuracy.
    VGG16 trained on CIFAR100 with SGD.
    Number of RND labels $n=2$, so $50\%$ accuracy corresponds to no memorization.
    \textbf{B+C:} While the regularization effectively reduces memorization, especially when no weight decay is used, there is no improvement in test accuracy.
    }
    \label{fig:copyDepth__reg}
\end{figure}

In the other sections of this paper, random prediction heads are applied only before the last fully connected layer.
However, these heads can also be constructed from earlier layers.
In this section, we investigate the effect of placing the random prediction heads at various depths within the network.
We therefore introduce the copy depth parameter $d$, where the value $d=1$ corresponds to copying only the last layer, as done in the rest of this paper.
To allow copying layers at any position, in this section we investigate a network without residual connections, in particular a VGG16~\citep{VGG}, which consists of 16 sequential layers: the first 13 are convolutional, followed by 3 fully connected layers.
The additional random label prediction heads copy all layers from the chosen copy location onwards.
Thus, $d=16$ corresponds to copying the entire network for each head.
The results are shown in \figrefA{fig:copyDepth__reg}.
Conventionally, CNNs are viewed as feature extractors in their convolutional stages, transforming inputs into abstract feature maps.
Although abstract, these features remain sample-specific.
The subsequent fully connected layers then transform these features into class labels.
Interestingly, while the analysis shows a steep increase in random label prediction accuracy, this increase does not occur at the transition between the convolutional and fully connected layers (i.e., between layers 3 and 4, counted from the network output).
Instead, accuracy increases between layers 5 and 7, which is still within the convolutional part of the network.
Since learning random labels requires sample-specific information, random label accuracy can be seen as an indicator of the transition from sample-specific to class-specific information within the feature maps.
These results suggest a clear transition between feature extraction and classification in the network, but this transition does not occur between the fully connected and convolutional layers, but rather deeper within the convolutional part of the network.\\\
An alternative explanation for the increase in random label accuracy could be the increasing capacity of the random prediction heads, which grows as more layers are copied.
To test this hypothesis, we increased the width of the copied layers at a copy depth of $d=3$, but did not observe a significant increase in memorization.

\subsection{Regularization}
The effect of applying the proposed regularization techniques, i.e. setting $\lambda > 0$, is examined in \figrefBC{fig:copyDepth__reg}.
The regularization successfully reduces the random label prediction accuracy, thus, mitigating memorization.
However, no improvement in test accuracy is observed, contradicting the expectation that reduced memorization would lead to better generalization, as suggested by generalization bounds based on Rademacher complexity.

\section{Conclusion}
This study examined the relationship between memorization, quantified as random label prediction accuracy, and several complexity regularizers, namely dropout, weight decay, and label smoothing.
The findings indicate that the proposed memorization metric serves as a valid measure of model complexity, consistent with the theoretical motivation of the proposed method based on Rademacher complexity.
Additionally, the introduced metric provided valuable insights into the transition from sample-specific image data to class-specific information, offering new perspectives on feature extraction and classification in CNNs.\\
However, when the proposed memorization metric was applied as an additional regularizer, the results differed from theoretical expectations.
Although the metric effectively reduced memorization, it did not lead to improved generalization.
This outcome challenges predictions from PAC learning theory and Rademacher complexity.\\
There are several limitations to this study.
It focuses primarily on CNNs and image classification tasks, specifically the CIFAR100 dataset.
Due to its multi-head structure, the proposed architecture does not scale efficiently to tasks with many classes.
Furthermore, the proposed architecture only measures memorization at the layer where additional prediction heads are attached.
Hence, potential shifts in memorization to earlier layers, induced by regularization, rather than eliminating memorization altogether, remain undetected.\\
In future work we plan to further analyze the impact of the proposed regularizer on the network, specifically investigating whether memorization shifts occur and how these shifts can be mitigated.
Independent of our technical proposal, we are interested in alternative realizations of the general concept of random label regularization to mitigate overfitting.\\
The appendix offers further analysis of the relation to the learning rate and label smoothing as well as a simplified network architecture that eliminates the need for the hyperparameter $\lambda$.
\newpage

\section*{Acknowledgements}
This work was funded by the Deutsche Forschungsgemeinschaft (DFG CRC 1459 Intelligent Matter Project-ID 433682494).

\bibliography{References}
\bibliographystyle{iclr2024_conference}

\newpage
\appendix

\makeatletter
\renewcommand{\@seccntformat}[1]{}
\makeatother
\section{Appendix}
\makeatletter
\renewcommand{\@seccntformat}[1]{\csname the#1\endcsname\quad}
\makeatother
\renewcommand\thefigure{\thesection.\arabic{figure}}
\setcounter{figure}{0}
\renewcommand\thetable{\thesection.\arabic{table}}
\setcounter{table}{0}

\subsection{Related Work}

The phenomenon of data memorization, although not new, gained significant attention in the era of modern deep learning with the works of \citet{zhang_understanding_2017} and \citet{arpit_closer_2017}.
Further studies on memorization have been conducted by e.g. \citet{NEURIPS2020_1e14bfe2,memorizationNNs,NEURIPS2019_dbea3d0e}, with a special focus on the effects of heavy overparameterization by \cite{memOver}, as well as with minimal overparameterization by \cite{memNOover}.
While the work presented here focuses on memorization in vision models, language models often show even more pronounced memorization, often leading to privacy risks if training data can be extracted from the models, as highlighted by \citet{mem_language1} and \citet{mem_language2}.\\
Efforts to mitigate generalization and thus data memorization have included regularization techniques such as dropout \citep{dropout} and weight decay \citep{weightDecay}.
However, to the best of our knowledge, no approach to directly regularize memorization, as presented in this work, has been previously published.\\
Similar to the desired reduction of single-sample specific features, the suppression of unwanted features is a key concern in fair AI to avoid biased models caused by biases in the training data itself, e.g., in the form of gender, ethnicity, or religion \citep{fairAIsurvey,tian_image_2022,wang_towards_2020,zhang_mitigating_2018}.
The network structure proposed here has similarities to the multi-head structure proposed by \citet{alvi_turning_2018} in fair AI, though our proposed architecture uses a different head structure as well as a different loss formulations.\\
Additionally, interpreting the random label accuracy as a measure of information abstraction shares conceptual similarities with mutual information, as applied to ANNs by \citet{mutualInfo}.
Moreover, interpreting random label accuracy as a measure of information abstraction has conceptual similarities to mutual information as applied to ANNs by \citet{mutualInfo}.

\subsection{Rademacher Complexities}
\label{sec:rademacher}
Rademacher complexities quantify the ability of a model to fit randomly assigned labels. 
In the context of binary classification, this can be formally defined as follows:\\
\textbf{Rademacher Complexity for Binary Classification~\citep{foundationsML}:} Given a hypothesis class $\mathcal{H}$, train data $\{x_1,...,x_n\}$, i.i.d. uniform random variables $\sigma_1,...,\sigma_n \in \{\pm1\}$:
\begin{equation}
    \label{eq:rademacher_compl}
    \mathfrak{R}_n (\mathcal{H}) = \mathbb{E}_\sigma \bigg[\sup_{h \in \mathcal{H}} \frac{1}{n} \sum_{i=1}^{n} \sigma_i h(x_i) \bigg]
\end{equation}
In binary classification, the model's alignment with the true labels can be evaluated by the product of the labels and the model's output.
The model $h$ is chosen as a supremum over the model class, which in practice can be approximated by empirical risk minimization.
However, this supremum makes an exact evaluation of Rademacher complexity difficult.\\
Rademacher complexity is model-agnostic, meaning that it does not depend directly on specific attributes of the model, such as the number of parameters or its architecture.
Instead, it evaluates the model class based on its ability to fit the data.
In PAC-learning Rademacher complexity can be used to derive bounds on the generalization error of particular model classes.
Specifically, in binary classification, an upper bound on the generalization error is provided as:
\begin{thm}
  \label{eq:PAC_rademacher}
    Given a hypothesis class $\mathcal{H}$, train data $\mathcal{S} = \{(x_1,\sigma_1),...,(x_n,\sigma_n)\}$, with $\sigma_1,...,\sigma_n \in \{\pm1\}$, then for any $\delta > 0$, with probability at least $1-\delta$ for any $h \in \mathcal{H}$ it holds, that
    \[ R(h) \leq \hat{R}_\mathcal{S} (h) + \mathfrak{R}_n (\mathcal{H}) + \sqrt{\frac{\log(1/\delta)}{2n}}. \]
\end{thm}
Where $\hat{R}_\mathcal{S} (h)$ is the empirical error on the training dataset \citep{foundationsML}.

\subsection{Implicit Effect on Learning Rate Finetuning}

\begin{figure}[htb]
  \centering
  \begin{subfigure}[b]{0.5\textwidth}
      \centering
      \includegraphics[width=\textwidth]{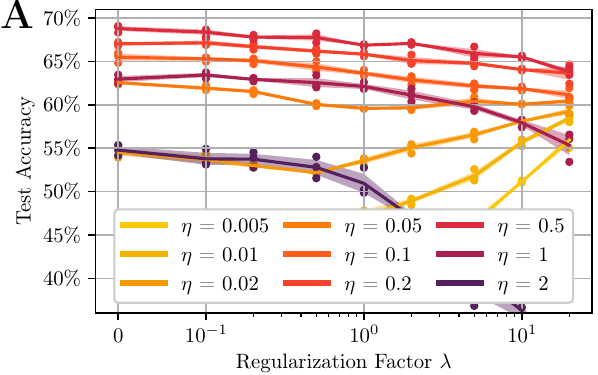}
    \end{subfigure}%
    \begin{subfigure}[b]{0.5\textwidth}
      \centering
      \includegraphics[width=\textwidth]{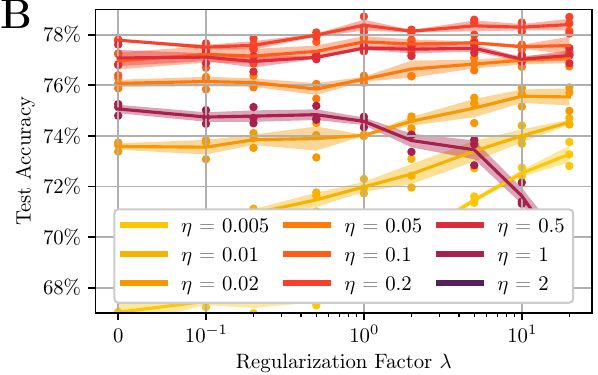}
    \end{subfigure}
  \caption{
    Dependence of regularization factor $\lambda$ and learning rate $\eta$.
    WideResNet-16-4 trained on CIFAR100 with number of random labels $n=10$.
    \textbf{A:} No augmentation. \textbf{B:} Including flipping, cropping and cutout.
    If the learning rate is too low, increasing $\lambda$ may have a positive effect, which could be misinterpreted as a reduction in memorization, but is actually caused by implicit learning rate fine-tuning of the regularizer.
    }
   \label{fig:RLR:LR_impact}
\end{figure}

The proposed regularization not only facilitates the unlearning of the random label but also trains the feature extractor to generate discriminative features for class label prediction.
This results from the fact that the summation in the definition of $L^{reg}$ is limited to the correct class head.
Thus, $L^{reg}$ has a positive training effect on the feature extractor (see \autoref{sec:RLR:singleOut} for a version of the proposed algorithm that is solely trained by this effect).
Therefore, the regularization parameter $\lambda$ has a significant impact on the learning rate, so it is necessary to carefully tune the learning rate before adding the regularizer.
However, this same effect makes $\lambda$ a stable hyperparameter, meaning that too high values of $\lambda$ do not cause dramatic performance drops, unlike many other regularizers.
As shown in \autoref{fig:RLR:LR_impact}, the regularization parameter $\lambda$ has a similar effect as increasing the learning rate $\eta$.
For high learning rates ($\eta > 0.5$), increasing $\lambda$ degrades performance.
However, with lower learning rates, the implicit learning effect of the regularizer compensates this, resulting in significant performance improvements as $\lambda$ is increased.
It is crucial to separate this interaction from the actual effect of the regularization itself.
Otherwise, any observed benefits may simply be due to implicit fine-tuning of the learning rate rather than to the mitigation of memorization.
In the experiment shown here, the regularization has a positive effect only for the optimal learning rate when augmentation is used (see \figrefB{fig:RLR:LR_impact}).
Otherwise, any observed benefits may simply be due to implicit fine-tuning of the learning rate rather than to the mitigation of memorization.

\subsection{Relation to Label Smoothing}
\begin{figure}[htb]
  \centering
    \begin{subfigure}[b]{0.5\textwidth}
      \centering
			\includegraphics[width=\textwidth]{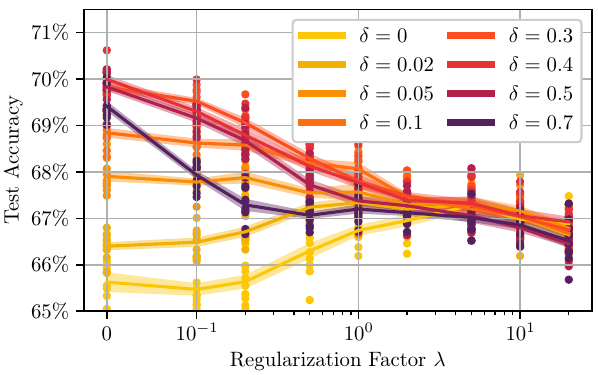}
    \end{subfigure}
		\caption{
      WRN16-4 trained on CIFAR100 with number of random labels $n=100$.
      The proposed random prediction regularizer improves the test accuracy only when low label smoothing is chosen.
      }
      \label{fig:RLR:reg_LS}
\end{figure}

Label smoothing \citep{labelSmoothing} alters the loss by evaluating the cross entropy against soft labels instead of one-hot encoded label vectors.
The resulting cross-entropy loss 
\begin{equation}
  L^{LS} = - (1-\delta) \log(p_s) - \sum_{\substack{i=1 \\ i\neq s}}^{N} \frac{\delta}{N-1} \log(p_i)
\end{equation}
can be reformulated to
\begin{equation}
  \label{eq:RLR:LS_uniform}
  L^{LS} = - \bigg(1-\delta-\frac{\delta}{N-1}\bigg) \log(p_s) - \sum_{i=1}^{N} \frac{\delta}{N-1} \log(p_i),
\end{equation}
with prediction output $p$, correct label index $s$, and label smoothing factor $\delta$.
Setting $\delta=0$ restores the default cross entropy loss for one-hot encoded targets, i.e.,
\begin{equation}
  L^{LS} = - \log(p_s)
\end{equation}
From the formulation in equation \autoref{eq:RLR:LS_uniform} it is apparent that label smoothing combines a rescaling of the learning rate (which is equivalent to a rescaling of the loss function) with an additional loss term that regularizes the model towards a uniform distribution of the predictions.
Both effects are also implicit in the proposed regularization technique.
However, unlike label smoothing, the proposed regularizer does not impact the actual predictions used for classification.
Instead, it is generated from the additional random label prediction heads and only affects the network from the last feature layer backwards.
In a comprehensive study of the effects of label smoothing, \citet{when_LS_helps} suggested that the main effect of label smoothing occurs in the penultimate layers, which in the case of the WideResNet architecture primarily studied here, corresponds to the last feature layer.
Therefore, we expect similar effects between label smoothing and the proposed regularizer.\\
\autoref{fig:RLR:reg_LS} shows the test accuracies for various combinations of label smoothing factor $\delta$ and random label regularization factor $\lambda$.
Without random label regularization (i.e., $\lambda = 0$), label smoothing achieves its highest performance improvements of more than $4\%$ for $\delta = 0.3$.
With little or no label smoothing ($\delta = 0$), increasing $\lambda$ leads to significant accuracy gains.
However, at larger values of $\delta$, including the optimal $\delta = 0.3$, the proposed regularization leads to a monotonic decrease in test accuracy.
For high regularization factors $\lambda$, label smoothing has no significant effect, either positive or negative.\\
Combined with the empirical results from the previous section, these findings challenge the hypothesis that the proposed regularizer improves generalization by reducing sample memorization.
Instead, the results support an alternative hypothesis: the proposed regularizer exhibits an implicit effect similar to label smoothing.
While this effect improves performance in the absence of label smoothing, it provides no additional benefit when optimal label smoothing is already applied.
The proposed regularizer does not achieve test accuracies comparable to those of label smoothing.
Moreover, for large regularization factors $\lambda$, the same results are observed for all label smoothing factors $\delta$.
This suggests that in addition to the implicit effect of the regularizer, which is similar to label smoothing, there is a second, detrimental effect that counteracts label smoothing.
As $\lambda$ increases, this second effect dominates, negating the benefits of label smoothing for large $\lambda$ values.\\

\subsection{Single Output Variant}
\label{sec:RLR:singleOut}
\begin{figure}[htb]
  \centering
  \includegraphics[width=0.5\textwidth]{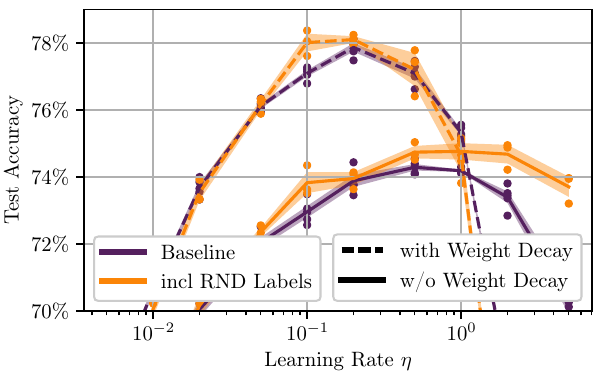}
  \caption{
    Single prediction head architecture.
    WRN16-4 trained on CIFAR100 with number of random labels $n=100$.
    Used augmentations: flipping, cropping, and cutout.
    Only small improvements in test accuracy occur.
  }
  \label{fig:RLR:singleOut}
\end{figure}
As previously discussed, the regularization loss also promotes learning of the correct class labels.
Since it is formulated to enforce an equal distribution of random predictions only in the correct class head, the head selection results in learning of the correct class.
Summing over the random predictions in each head allows generating class predictions from the random prediction heads:
\begin{equation}
  \hat{p}^{\text{class}}_j = \sum_{i}^{n} \hat{p}_i^j
\end{equation}
This eliminates the need for an additional class prediction layer.
As a result, the feature extractor can be trained using only the regularization loss $L^{reg}$, while the random prediction heads are still trained by the random label loss $L^{rnd}$.
Consequently, the class prediction loss $L^{class}$ and the additional hyperparameter $\lambda$ are no longer needed.
While this simplifies the model by removing the need to tune an additional parameter, it also limits the ability to continuously control the regularization effect.\\
Nevertheless, this variant of the random label learning architecture allows learning (and thus memorization) of the random labels.
Experimental results shown in \autoref{fig:RLR:singleOut} demonstrate that only marginal improvements in test accuracy can be achieved both with and without weight decay.
Furthermore, for several other hyperparameter settings, such as training without augmentation, there were significant decreases in test accuracy.

\end{document}